\documentclass[10pt,twocolumn,letterpaper]{article}

\usepackage{cvpr}              

\usepackage[dvipsnames]{xcolor}

\definecolor{cvprblue}{rgb}{0.21,0.49,0.74}
\usepackage[pagebackref,breaklinks,colorlinks,citecolor=cvprblue]{hyperref}

\usepackage[accsupp]{axessibility} 

\usepackage{bm}
\usepackage{capt-of}
\usepackage{mathtools}
\usepackage{siunitx}
\usepackage[figurename=Figure]{caption}
\usepackage{scrextend}
\usepackage{xspace}
\usepackage{xcolor}
\usepackage{multirow}
\usepackage{wrapfig}
\usepackage{amsthm}
\usepackage{thmtools}
\usepackage{algorithm,algpseudocode}
\usepackage[export]{adjustbox}
\usepackage{url}
\usepackage{lineno}
\usepackage{pifont}
\usepackage{booktabs}
\usepackage{soul}

\newcounter{algsubstate}

\newtheorem*{theorem*}{Theorem}

\renewcommand{\eqref}[1]{\mbox{Eq.~(\ref{#1})}}

\makeatletter
\DeclareRobustCommand\onedot{\futurelet\@let@token\@onedot}
\def\@onedot{\ifx\@let@token.\else.\null\fi\xspace}
\def\eg{\emph{e.g}\onedot}
\def\ie{\emph{i.e}\onedot}

\makeatother

\newcolumntype{L}[1]{>{\raggedright\arraybackslash}p{#1}}
\newcolumntype{C}[1]{>{\centering\arraybackslash}p{#1}}
\newcolumntype{R}[1]{>{\raggedleft\arraybackslash}p{#1}}

\newcommand{\cmark}{\ding{51}}%
\newcommand{\xmark}{\ding{55}}%

\newcommand*\samethanks[1][\value{footnote}]{\footnotemark[#1]}
\newcommand\correspondingauthor{\thanks{Contributed equally as co-senior authors.}}

\def\paperID{10816} 
\def\confName{CVPR}
\def\confYear{2024}

\title{FairCLIP: Harnessing Fairness in Vision-Language Learning}

\author{Yan Luo\textsuperscript{1}\thanks{Contributed equally as co-first authors.} \quad Min Shi\textsuperscript{1}\samethanks \quad Muhammad Osama Khan\textsuperscript{2}\samethanks \quad Muhammad Muneeb Afzal\textsuperscript{2} \quad Hao Huang\textsuperscript{3} \\ Shuaihang Yuan\textsuperscript{3} \quad Yu Tian\textsuperscript{1} \quad Luo Song\textsuperscript{1} \quad Ava Kouhana\textsuperscript{1} \quad Tobias Elze\textsuperscript{1} \quad Yi Fang\textsuperscript{2,3}\correspondingauthor \quad Mengyu Wang\textsuperscript{1}\textsuperscript{$\dagger$} \\
\textsuperscript{1}Harvard Ophthalmology AI Lab, Harvard University\\
\textsuperscript{2}Tandon School of Engineering, New York University \\
\textsuperscript{3}Multimedia and Visual Computing Lab, New York University Abu Dhabi \\
{\tt\small \{yluo16, mshi6\}@meei.harvard.edu, \{osama.khan, muneeb.afzal, hh1811, sy2366\}@nyu.edu,
} \\
{\tt\small \{ytian11, lsong7, akouhana, tobias\_elze\}@meei.harvard.edu,
} \\
{\tt\small yfang@nyu.edu, mengyu\_wang@meei.harvard.edu
}
}

\begin{document}
\maketitle
\begin{abstract}
Fairness is a critical concern in deep learning, especially in healthcare, where these models influence diagnoses and treatment decisions. Although fairness has been investigated in the vision-only domain, the fairness of medical vision-language (VL) models remains unexplored due to the scarcity of medical VL datasets for studying fairness. To bridge this research gap, we introduce the first fair vision-language medical dataset (\textit{Harvard-FairVLMed}) that provides detailed demographic attributes, ground-truth labels, and clinical notes to facilitate an in-depth examination of fairness within VL foundation models. Using Harvard-FairVLMed, we conduct a comprehensive fairness analysis of two widely-used VL models (CLIP and BLIP2), pre-trained on both natural and medical domains, across four different protected attributes. Our results highlight significant biases in all VL models, with Asian, Male, Non-Hispanic, and Spanish being the preferred subgroups across the protected attributes of race, gender, ethnicity, and language, respectively. In order to alleviate these biases, we propose FairCLIP, an optimal-transport-based approach that achieves a favorable trade-off between performance and fairness by reducing the Sinkhorn distance between the overall sample distribution and the distributions corresponding to each demographic group. As the first VL dataset of its kind, Harvard-FairVLMed holds the potential to catalyze advancements in the development of machine learning models that are both ethically aware and clinically effective. Our dataset and code are available at \url{https://ophai.hms.harvard.edu/datasets/harvard-fairvlmed10k}.
\end{abstract}    
\section{Introduction}
\label{sec:intro}
Fairness has received increasing interest in deep learning in recent years. This is vital, especially in healthcare, where these deep learning models influence diagnoses and treatment decisions. Biases in these models related to factors like race, gender, or socioeconomic status can lead to healthcare disparities and adverse patient outcomes. Hence, ensuring that these models are free from bias is not only an ethical and legal requirement but also a necessity for ensuring patient safety and healthcare equity. This makes fairness in medical computer vision a critical and immediate concern, essential for providing equitable healthcare.

\begin{table}[!t]
	\centering
	\caption{\label{tbl:dataset}
	Public medical fairness datasets. In contrast to existing medical VL datasets, our Harvard-FairVLMed provides detailed demographic attributes, ground-truth labels, and clinical notes with imaging as well as non-imaging clinical information.  
	}
        \vspace{-1ex}
	\adjustbox{width=1\columnwidth}{
	\begin{tabular}{ L{18ex} L{32ex} L{12ex} L{12ex} L{18ex} L{2ex} L{2ex} L{15ex}}
		\toprule
		\textbf{Dataset} & \textbf{Data Modality} & \textbf{\# of Images} & \textbf{\# of Patients}  & \textbf{Identity Attribute} & \textbf{V} & \textbf{L} & \textbf{L Info Type}\\

\cmidrule(lr){1-1} \cmidrule(lr){2-2} \cmidrule(lr){3-3} \cmidrule(lr){4-4} \cmidrule(lr){5-5} \cmidrule(lr){6-6} \cmidrule(lr){7-7} \cmidrule(lr){8-8}

Fitzpatrick17k \cite{groh2021evaluating}  & Skin Photos  & 16,012  & 1,373 & Skin type  & \cmark & \xmark\\

HAM10000 \cite{tschandl2018ham10000} & Dermatoscopy & 9,948  & - & Age; Gender & \cmark & \xmark\\

OL3I\cite{zambrano2021opportunistic} & Heart CT & 8,139   & 1,686 & Age; Gender  & \cmark & \xmark\\
ODIR-2019 \cite{odir2019} & Color Fundus & 8,000  & 5,000  & Age; Gender  & \cmark & \xmark\\
PAPILA \cite{kovalyk2022papila} & Color Fundus & 488  &  244 & Age; Gender  & \cmark & \xmark\\

Harvard-GDP \cite{Luo_2023_ICCV} & OCT & 1,000 & 1,000 & Age; Gender; Race; Ethnicity & \cmark & \xmark\\
COVID-CT-MD\cite{afshar2021covid} & Lung CT & 308 & 305 & Age; Gender & \cmark & \xmark\\

AMD-OCT\cite{farsiu2014quantitative} & OCT & 384 & - & Age  & \cmark & \xmark\\

ADNI 1.5T \cite{wyman2013standardization} & Brain MRI & 550 & & Age; Gender & \cmark & \xmark \\

   \cmidrule(lr){1-8}

   CheXpert \cite{irvin2019chexpert} & Chest X-ray and Radiology Report & 222,793   & 65,240 & Age; Gender; Race & \cmark & \cmark & Image Description\\

MIMIC-CXR \cite{Johnson_arXiv_2019} & Chest X-ray and Radiology Report & 370,955  &  65,079 & Age; Gender; Race & \cmark & \cmark & Image Description \\
PadChest \cite{bustos2020padchest} & Chest X-ray and Radiology Report (Spanish) & 160,868  & 69,882 & Age; Gender & \cmark & \cmark & Image Description\\

Harvard-FairVLMed & SLO Fundus and Clinical Note & Fundus: 10,000; Note: 10,000  & 10,000 & Age; Gender; Race; Ethnicity; Preferred Language; Marital Status & \cmark & \cmark & Image Description + Non-Imaging Clinical Info\\

		\bottomrule	
	\end{tabular}}
 \vspace{-7 mm}
\end{table}

Previous research has identified biases in deep learning-based medical imaging models, focusing primarily on Chest X-ray diagnosis~\cite{glocker2023algorithmic,khan2023fair}. In contrast to these vision-only models, there has been a recent surge of vision-language (VL) foundation models~\cite{luo2022biogpt,wang2022medclip,huang2023visual,xu2023elixr,kelly2024visiongpt,kelly2024visiongpt3d,yang2024worldgpt}, which have set new benchmarks across a wide spectrum of tasks. However, despite these strong performances, the fairness of these VL models remains unclear. Given the existence of biases in vision-only models (which operate on machine-acquired images) and the human-written nature of clinical notes, VL models could further aggravate fairness issues. Hence, as the field moves towards multi-modal foundation models, it becomes increasingly critical to scrutinize how the interplay of vision and text affects equity in algorithmic outcomes. The current landscape for such an investigation is, however, constrained by the scarcity of VL datasets that include comprehensive demographic information, with the existing public VL datasets primarily focused on Chest X-rays~\cite{irvin2019chexpert,johnson2019mimic}. Prior studies~\cite{seyyed2020chexclusion,irvin2019chexpert} have highlighted the challenges of studying fairness using these datasets since their \textit{ground-truth} labels are automatically extracted from radiology reports, potentially leading to inaccurate fairness conclusions due to the noisy labels. Moreover, since these datasets were not primarily designed for fairness, they provide only a handful of demographic characteristics, limiting the potential for a comprehensive fairness study across multiple dimensions. Furthermore, radiology reports, focusing mainly on direct observations of the imaging data with limited additional patient-specific information, are not representative of most clinical text, thus limiting their utility in fairness studies of medical VL models.

To bridge this research gap, we introduce the first fair vision-language medical dataset (\textit{Harvard-FairVLMed} for short) that provides detailed demographic attributes, ground-truth labels, and clinical notes to facilitate an in-depth examination of fairness within VL foundation models (Table \ref{tbl:dataset}). Harvard-FairVLMed contains records for 10,000 patients, each paired with an SLO fundus image and a clinical note for diagnosing Glaucoma, along with fine-grained protected attributes such as age, gender, race, ethnicity, preferred language, and marital status. Unlike radiology reports, the clinical notes in our dataset provide much more detailed information featuring not only image descriptions but also rich non-imaging clinical information such as medication, non-imaging test results, and family history. Hence, these clinical notes are more representative of clinical textual information, making them better suited for studying the fairness of medical VL models.

Glaucoma, affecting millions globally~\cite{gupta2016prevalence,tham2014global,tian2023fairseg, luo2023eye,shi2023jbhi, shi2024rnflt2vec, luo2023harvard,shi2023artifact}, exemplifies the need for equitable diagnostic models. Timely detection is crucial in order to avoid irreversible vision loss. However, many patients remain undiagnosed~\cite{shaikh2014burden} due to the disease's asymptomatic nature and barriers to ophthalmic care. Moreover, this undiagnosed issue is even more pronounced among minority groups. For example, previous studies have shown that individuals from Black communities are 4.4 times more likely to have undiagnosed and untreated Glaucoma compared to their White counterparts~\cite{shaikh2014burden}, highlighting the importance of addressing healthcare disparities. Deep learning systems hold significant potential for improving healthcare. However, it is imperative to address potential fairness issues in these deep learning systems prior to their clinical implementation to ensure equitable healthcare delivery.

In this work, we conduct an extensive fairness analysis with two widely-used VL methods (\ie, CLIP \cite{radford2021learning} and BLIP2 \cite{li2023blip}) on Harvard-FairVLMed. Our experimental findings reveal significant disparities across various groups based on race, gender, ethnicity, and language. To address these fairness issues, we introduce FairCLIP, an optimal transport-based approach. FairCLIP is designed to enhance fairness by optimizing the Sinkhorn distance, thereby aligning the overall sample distribution with the distributions specific to each demographic group.




Our main contributions can be summarized as follows:
\begin{itemize}
    \item We introduce the first fair vision-language medical dataset (Harvard-FairVLMed) featuring detailed demographic attributes, ground-truth labels, and clinical notes for studying the fairness of VL foundation models.
    \item Using Harvard-FairVLMed, we conduct a comprehensive fairness analysis of two widely-used VL models (\ie, CLIP and BLIP2), pre-trained on both natural and medical domains, across four different protected attributes.
    \item Our results highlight significant biases in all VL models, with Asian, Male, Non-Hispanic, and Spanish being the preferred subgroups across the protected attributes of race, gender, ethnicity, and language, respectively.
    \item We propose an optimal transport-based approach named FairCLIP, which demonstrates a significant improvement over CLIP in terms of both performance and fairness.
\end{itemize}

\section{Related Work}
\label{sec:related_work}

\noindent\textbf{Fairness Models}
Fairness in machine learning is crucial for creating dependable systems that avoid discrimination. Historical instances of bias have been identified in various technologies. In the field of image processing, biases exist in facial recognition \cite{wang2020mitigating,wang2020towards,hanna2020towards,benthall2019racial,buolamwini2018gender} and pedestrian detection algorithms \cite{brandao2019age,li2020learning,kogure2022age}. Recent efforts in deep learning aim to rectify these biases by adjusting training processes to minimize unfair predictions across different demographic groups. These deep learning methods focus on removing or neutralizing sensitive features and fostering representations that are fair and unbiased. 

Particularly in the medical domain, fairness models have emerged as a critical tool to mitigate biases and promote equity, particularly within healthcare applications. This field has developed a range of methodologies, each targeting a different stage: pre-processing, in-processing, and post-processing. 
Pre-processing strategies ~\cite{puyol2021fairness,brown2022detecting,seyyed2020chexclusion,zhou2021radfusion,joshi2021ai,ramaswamy2021fair,zhang2020towards,park2022fair,ramaswamy2021fair,zhang2020towards,zietlow2022leveling,tian2023fairseg} are designed to mitigate biases by enhancing the dataset prior to model training.
In in-processing approaches ~\cite{beutel2017data,roh2020fr,sarhan2020fairness,zafar2017fairness,zhang2018mitigating,beutel2017data,roh2020fr,zhang2018mitigating,sarhan2020fairness,shi2023equitable} fairness is incorporated directly into the model’s training stage. 
Post-processing methods~\cite{pleiss2017fairness,marcinkevics2022debiasing,wu2022fairprune}, applied after the model has been trained, aim to adjust outputs to improve fairness.

To address the challenges of fairness in VL tasks, this work introduces FairCLIP, a method designed to promote fairness. FairCLIP aligns the distribution of all samples with those of specific demographic groups, leveraging the similarities between visual and textual features.
\noindent\textbf{Vision-Language Models}
The integration of vision and language in machine learning, especially through models like CLIP (Contrastive Language-Image Pretraining) \cite{radford2021learning}, represents a significant advancement in computer vision. CLIP, trained from a large scale of images and their related text descriptions, marks an important step in aligning visual information with text. BLIP (Bootstrapping Language-Image Pre-training) \cite{li2022blip} represents a novel approach in vision-language models, utilizing a visual transformer as an image encoder and a transformer for text processing. BLIP-2 \cite{li2023blip} advances the field with its scalable multimodal pre-training approach, enabling LLMs to process and comprehend images. 

The success of CLIP-like models has inspired numerous adaptations in the medical field \cite{zhang2023large,you2023cxr,liu2023clip,wu2023medklip,lei2023clip,wang2022medclip,zhou2023anomalyclip}, where it confronts challenges like limited data and the need for accurate interpretations.
For instance, Pathology Language–Image Pretraining (PLIP) \cite{huang2023visual} utilizes the OpenPath dataset \cite{huang2023visual}, consisting of pathology images with language descriptions derived from medical Twitter posts. 
PMC-CLIP (PubMedCentral-CLIP) \cite{lin2023pmc} joins these efforts by enhancing CLIP’s application in the biomedical domain.

Beyond these adaptations of CLIP, several other vision-language models have emerged to address various medical tasks that range from generating reports from medical images \cite{liu2019clinically,niksaz2023improving,nguyen2022eddie,yang2021automatic,wu2022agnet,tian2018diagnostic} to diagnosing medical conditions \cite{naseem2022vision,zhang2019pathologist,monajatipoor2022berthop,zhao2021automatically}, and facilitating medical image-based question-answering (VQA) \cite{bazi2023vision,kohli2022dermatobot}. 

Despite these advancements, a gap remains in understanding and addressing fairness concerns within vision-language models (VLMs). The complex interplay of visual and textual data in VLMs poses unique challenges for fairness, which have been largely unexplored, especially in the context of medical applications.

\noindent\textbf{Vision-Language Medical Fairness Datasets}
Medical fairness datasets serve a critical role in the development of machine learning models \cite{Irvin_AAAI_2019,Johnson_arXiv_2019,Luo_2023_ICCV,luo2023eye,Luo_arXiv_2023}, ensuring that they operate equitably across diverse patient populations. In this context, datasets equipped with imaging data as well as associated textual reports are particularly valuable. They not only enable the development of models but also allow for nuanced explorations of fairness by providing a richer context for each medical case.

CheXpert~\cite{Irvin_AAAI_2019} is a prominent chest radiography dataset annotated for 14 distinct observations. CheXpert uses an automated system to extract labels from radiology reports. However, this method introduces uncertainties due to the inherent ambiguities in the reports and varying interpretations by radiologists. This uncertainty, such as the 16\% of Atelectasis labels that are uncertain, represents a significant challenge in using datasets like this in real-world scenarios~\cite{Irvin_AAAI_2019,abdalla2023hurdles}. MIMIC-CXR~\cite{Johnson_arXiv_2019} contributes to this area in which each study is complemented by a semi-structured free-text radiology report. Padchest~\cite{bustos2020padchest} further expands the resources by providing chest X-ray images along with a set of radiographic findings that have been processed using radiology reports. However, the radiological reports are only available in Spanish, which may present a barrier for the global research community.
Similarly, datasets pertaining to image captioning~\cite{subramanian2020medicat,demner2016preparing,jing2017automatic,pelka2018radiology} and VQA ~\cite{liu2021slake,he2020pathological,lau2018dataset} have been developed in the medical domain. 

In contrast, fundus image datasets present a different scenario. While ODIR2019~\cite{odir2019} and PAPILA \cite{kovalyk2022papila} offer valuable imaging data for ocular conditions, they do not provide the same depth of textual information as CheXpert and MIMIC-CXR. Also, the Glaucoma Detection and Progression \cite{Luo_2023_ICCV}, Eye Fairness \cite{luo2023eye}, and Glaucoma Fairness \cite{Luo_arXiv_2023} datasets lack textual information required for studying the VL problem.
While datasets such as CheXpert and MIMIC-CXR have advanced the field of fairness research by pairing images with radiology reports, they fall short in providing extensive demographic data and are prone to noisy labels due to their automated labeling processes. Additionally, they were not initially created with a focus on fairness, which poses limitations in studying fairness.

Our Harvard-FairVLMed dataset stands as a pioneering resource tailored for fairness studies in VL modeling. It comprises ground-truth labels, ensuring high label quality, and includes a broad range of identity attributes such as age, gender, race, ethnicity, preferred language, and marital status. Hence, this dataset not only enhances the quality and reliability of fairness studies but also broadens the horizon for research in the VL domain.

\section{Dataset Analysis}
This study strictly adheres to the principles outlined in the Declaration of Helsinki and has been approved by our institute's Institutional Review Board. All data in this dataset are de-identified.

\begin{figure}
  \centering
    \includegraphics[width=0.48\textwidth]{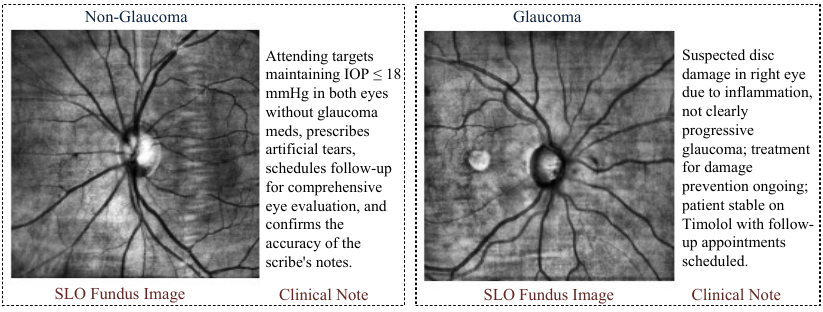}
     \vspace{-4mm}
  \caption{Examples of non-glaucomatous and glaucomatous samples with the corresponding SLO fundus image and clinical note.} 
  \label{fig:example}
\end{figure}

\subsection{Data Collection and Quality Control}

Harvard-FairVLMed contains subjects who received glaucoma care from Massachusetts Eye and Ear at Harvard Medical School between 2015 and 2022. There are three types of data to be released in this study: (1) scanning laser ophthalmoscopy (SLO) fundus images; (2) demographic identity group information; and (3) de-identified clinical notes written by ophthalmologists to provide a summary of the glaucoma diagnosis. SLO fundus images are a valuable marker for assessing retinal damage from diseases like glaucoma. Each SLO fundus image is associated with six demographic identity attributes, including age, gender, race, ethnicity, preferred language, and marital status. The accompanying clinical notes vary in length. These notes may detail assessments, treatment plans, and diagnostic strategies, and are considered to correspond with the visual semantics in SLO fundus images. Two examples of pairs of SLO fundus images and clinical notes are shown in Figure \ref{fig:example}. The subjects are categorized into non-glaucoma (visual function measured by visual field (VF) test is normal: VF mean deviation $\geq$ -1 dB and normal VF glaucoma hemifield test and pattern standard deviation (PSD) results) and glaucoma categories (visual function measured by VF test is abnormal: VF mean deviation $<$ -3 dB and abnormal VF glaucoma hemifield test and PSD results).

\subsection{Protected Information De-Identification}
The raw clinical notes may contain protected sensitive information, such as the date of glaucoma diagnosis, patient name, phone number, email address, physical location, institution, etc. We de-identified this sensitive information in the following three steps. First, we anonymized all the clinical notes using Microsoft’s Presidio \footnote{https://github.com/microsoft/presidio.}, which replaced sensitive information with respective placeholders (e.g., PERSON\_NAME, PHONE\_NUMBER, and LOCATION) so that the original sentence structure and coherence can be maintained. Then, we used rules to match and de-identify the protected information (e.g., physical address) that has not been fully recognized by Presidio. Lastly, the de-identified clinical notes were further verified by four medical experts. Specially, every clinical note was checked by an expert and the sensitive information was manually replaced with respective placeholders when necessary. 


\subsection{Data Characteristics}

Our Harvard-FairVLMed dataset comprises 10,000 samples from 10,000 subjects. It is divided into 7,000 training, 1,000 validation, and 2,000 test samples.
The dataset's collective age average is 60.9 $\pm$ 16.2 years. The dataset includes samples from three major groups: Asian, with 819 samples; Black, with 1,491 samples; and White, with 7,690 samples. Gender-wise, females constitute 56.3\% of the subjects, with the remainder being males. The ethnic distribution is highlighted by 90.6\% Non-Hispanic, 4.0\% Hispanic, and 5.4\% unspecified. In terms of preferred language, 92.5\% of the subjects prefer English, 1.7\% prefer Spanish, 0.8\% prefer other languages, and 5.0\% remain unknown. From a marital status perspective, 57.4\% are in a marriage or partnered, 26.4\% are single, 6.6\% have experienced divorce, 1.0\% are legally separated, 6.1\% are widowed, and 2.5\% are not specified. After de-identification, clinical notes vary from 11 to 332 words, with an average word count of 147.

\section{Methodology}

\subsection{Background}

\begin{figure*}[t]
  \centering
    \includegraphics[width=0.8\textwidth]{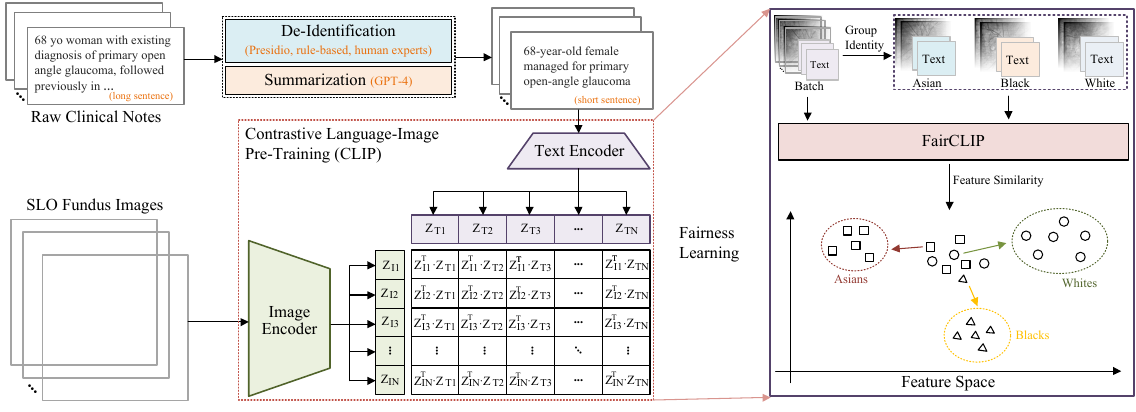}
  \caption{
  Schematic view of the proposed FairCLIP. Clinical notes containing PHI (\eg, names and gender) undergo de-identification and summarization to fit text encoder limitations, such as CLIP's 77-token maximum length. FairCLIP equalizes the overall sample distribution with the distributions corresponding to each demographic group, thereby achieving a favorable trade-off between performance and fairness.
  }
  \label{fig:overview}
\end{figure*}

With the labeled data $D=\left\{\left(x_{I i}, x_{T i}, y_i, a_i\right)\right\}$, where $x_{I i} \in \mathcal{X}_{I}$ represents an SLO fundus image, $x_{T i} \in \mathcal{X}_{T}$ represents a clinical note, $y \in \mathcal{Y}$ denotes a glaucoma diagnosis label, and $a \in \mathcal{A}$ signifies an identity attribute associated with the patient, such as gender, race, or ethnicity. To process multi-modal data, a vision-language (VL) pre-trained model (e.g., CLIP \cite{Radford_ICML_2021}) $f$ employs a vision encoder $f_I$ and a text encoder $f_T$ to generate vision features $z_I$ and text features $z_T$, respectively.
Given a batch of $N$ image and text pairs, CLIP is trained to generate a similarity matrix $M \in \mathbb{R}^{N \times N}$ with each entry $M_{ij} = z_{Ii}^\top z_{Tj}$ representing the cosine similarity between the $i$-th image feature $z_{Ii}$ and the $j$-th text feature within the given batch. To achieve this, CLIP adopts a contrastive learning scheme to maximize the cosine similarity of the image and text features of the $N$ positive pairs in the batch, while minimizing the cosine similarity of the image and text features of the remaining $N^2-N$ negative pairs. Following \cite{Radford_ICML_2021}, a symmetric cross-entropy loss is calculated over $M$ to optimize $f_I$ and $f_T$. Mathematically, the optimization goal is defined as follows:
$$
\min_{f} -\sum_{i=1}^N \sum_{j=1}^N \delta(i-j)\log \left(z_{I i}^{\top} z_{T j} /\left(\|z_{I i}\| \cdot\|z_{T j}\|\right)\right),
$$
where $\delta(i-j)$ is Dirac's Delta function, $\delta(i-j)=1$ when $i=j$.
In the context of fairness learning, it requires to incorporate identity information during model training, i.e., $f \in \mathcal{F}: \in \mathcal{X}_{I} \times \mathcal{X}_{T} \times \mathcal{A} \xrightarrow{\theta} \mathcal{Y}$. Consequently, fairness learning aims to minimize disparities between different identity groups while also maximizing accuracy.



\subsection{FairCLIP}


As illustrated in Figure \ref{fig:overview}, the proposed FairCLIP framework is designed to improve fairness during the pre-training phase. This is achieved by minimizing the disparity between the probability distributions of $M_{i,i}$, which represents the correlation between visual and language features, across different racial groups (or other attribute-based groups). Denote $\mathcal{D}_{\{(x_{I}, x_{T}, a)\}|f}$ as a distribution of $M_{i,i}$ given the model $f$. If the samples are from a specific group (\eg, white), the corresponding distribution is $\mathcal{D}_{\{(x_{I}, x_{T}, a)| a = \text{white}\}|f}$. Then, the objective of enhancing fairness can be defined as:
\begin{align}
    \begin{split}
        \min_{f} \sum_{\alpha}^{\mathcal{A}} d (\mathcal{D}_{\{(x_{I}, x_{T}, a)\}|f} - \mathcal{D}_{\{(x_{I}, x_{T}, a)| a = \alpha\}|f})
    \end{split}
    \label{eqn:fairclip_obj}
\end{align}
where $d$ is a distance function, $\mathcal{D}_{\{(x_{I}, x_{T}, a)\}|f}$ and $\mathcal{D}_{\{(x_{I}, x_{T}, a)| a = \alpha\}|f}$ are the underlying distributions that are computationally intractable. Instead, we use empirical distributions in Equation (\ref{eqn:fairclip_obj}) that are estimated based on batch $\mathcal{B}$, \ie, $\mathcal{D}_{\mathcal{B}|f}$ and $\mathcal{D}_{\mathcal{B}_{a}|f}$. $\mathcal{B}_{a}$ means the samples in the batch are from group $a$. 



To optimize the objective (\ref{eqn:fairclip_obj}), one straightforward way is to minimize Kullback–Leibler (KL) divergence between the two distributions. However, KL-divergence is not symmetric and does not satisfy the triangle inequality, and thus not a true metric of distance. Instead, we follow \cite{peyre_FTML_2019} to minimize the Sinkhorn distance between the two distributions. Sinkhorn distance is a type of probability metric and a variant of the Wasserstein distance.

Specifically, we have a measurable set $\mathcal{Z}=\{z^{\top}_{I}z_{T}\}$, denoted by $\mathcal{M}(\mathcal{Z})$ the set of measures (not necessarily probability measures) on $\mathcal{Z}$, and $\mathcal{P}(\mathcal{Z})$ the set of probability measures on $\mathcal{Z}$. 
Consider distributions $\mathcal{D}_{\mathcal{B}}, \mathcal{D}_{\mathcal{B}_{a}} \in \mathcal{P}(\mathcal{Z})$, and let $\mu, \nu \in \mathcal{M}(\mathcal{Z})$ be two reference measures.
For regularization parameter $\epsilon \geq 0$, the Sinkhorn distance between two distributions $\mathcal{D}_{\mathcal{B}}$ and $\mathcal{D}_{\mathcal{B}_{a}}$ is defined as
\begin{align}
    \begin{split}
        &\mathcal{W}_\epsilon(\mathcal{D}_{\mathcal{B}}, \mathcal{D}_{\mathcal{B}_{a}}) = \\ &\inf _{\gamma \in \Gamma(\mathcal{D}_{\mathcal{B}}, \mathcal{D}_{\mathcal{B}_{a}})}\left\{\mathbb{E}_{(x, y) \sim \gamma}[c(p, q)]+\epsilon H(\gamma \mid \mu \otimes \nu)\right\},
    \end{split}
\end{align}
where $\Gamma(\mathcal{D}_{\mathcal{B}}, \mathcal{D}_{\mathcal{B}_{a}})$ denotes the set of joint distributions whose first and second marginal distributions are $\mathcal{D}_{\mathcal{B}}$ and $\mathcal{D}_{\mathcal{B}_{a}}$, respectively, $c(p, q)$ denotes the transport cost, and $H(\gamma \mid \mu \otimes \nu)$ denotes the relative entropy of $\gamma$ with respect to the product measure $\mu \otimes \nu$. $p$ and $q$ are the points in the distributions $\mathcal{D}_{\mathcal{B}}$ and $\mathcal{D}_{\mathcal{B}_{a}}$, respectively. 
The Sinkhorn loss would be added to the loss used by CLIP in the pre-training phase for optimizing each component in CLIP.

\section{Experiment \& Analysis}

\subsection{Experimental Setup}
\label{sec:experimental_setup}
In this section, we describe the pre-training and evaluation setups as well as the metrics used to study performance and fairness of VL models on our proposed Harvard-FairVLMed dataset.

\noindent\textbf{Pre-Training:} We use the widely-adopted VL methods -- CLIP \cite{Radford_ICML_2021} and BLIP2 \cite{li2023blip} -- for our analysis. For the natural pre-trained variants, we use the official checkpoints provided by CLIP and BLIP2. For the medical pre-trained variants, we pre-train both methods on the Harvard-FairVLMed dataset after initializing from the official checkpoints. More details are in the supplementary material.

\noindent\textbf{Evaluation:} We utilize two types of evaluation strategies -- linear probing and zero-shot transfer. For linear probing, we follow the official MAE~\cite{he2022masked} implementation and train a linear classifier on top of the visual features from CLIP and BLIP2, respectively. Similar to MAE, we use a BatchNorm layer~\cite{ioffe2015batch} before the linear classifier and employ a LARS optimizer~\cite{you2017large} with a base learning rate of 0.1, weight decay of 0, and batch size of 512. We conduct linear probing for 1000 epochs on a single V100 GPU for all experiments.
For zero-shot transfer, we use the same setup as CLIP and select the class corresponding to the text embedding, which has the highest similarity with the image embedding.

\begin{table*}[t]
\centering
\scriptsize
\caption{VL fairness analysis of two widely-used VL methods, \ie, CLIP and BLIP2, pre-trained on both natural and medical domains and evaluated across four different protected attributes, with all scores presented in percentage.}
\vspace{-2ex}
\label{tab:vl_fairness_benchmark}
\adjustbox{width=.85\textwidth}{
\begin{tabular}{lllllllll}
\toprule
\textbf{Attribute} & \textbf{Model}  & \textbf{DPD $\downarrow$} & \textbf{DEOdds $\downarrow$} & \textbf{AUC $\uparrow$} & \textbf{ES-AUC $\uparrow$} & \multicolumn{3}{c}{\textbf{Group-wise AUC $\uparrow$}} \\ \midrule
&&&&&&\textbf{Asian} & \textbf{Black} & \textbf{White} \\
\multirow{4}{*}{\textbf{Race}} & CLIP & 5.30 ± 0.63 & 14.00 ± 1.01 & 77.27 ± 0.03 & 72.43 ± 0.29 & 79.74 ± 0.31 & 73.60 ± 0.12 & 77.82 ± 0.03 \\
                      & CLIP-FT & 4.01 ± 0.47 & 9.57 ± 0.83 & 80.27 ± 0.08 & 74.70 ± 0.33 & 82.19 ± 0.26 & 75.67 ± 0.21 & 81.20 ± 0.08 \\
                      & BLIP2 & 9.44 ± 0.65 & 10.62 ± 0.22 & 73.81 ± 0.02 & 68.88 ± 0.04 & 76.28 ± 0.06 & 69.55 ± 0.09 & 74.22 ± 0.03 \\
                      & BLIP2-FT & 8.30 ± 0.36 & 10.91 ± 0.32 & 80.10 ± 0.03 & 73.81 ± 0.10 & 82.09 ± 0.09 & 74.43 ± 0.08 & 80.97 ± 0.07 \\ \midrule
&&&&&&\textbf{Female} & \textbf{Male} \\
\multirow{4}{*}{\textbf{Gender}} & CLIP & 1.08 ± 0.19 & 5.19 ± 0.54 & 77.27 ± 0.03 & 72.47 ± 0.10 & 74.25 ± 0.07 & 80.88 ± 0.03 &  \\
                      & CLIP-FT & 0.39 ± 0.26 & 4.55 ± 0.33 & 80.27 ± 0.08 & 75.81 ± 0.12 & 77.59 ± 0.09 & 83.47 ± 0.04 &  \\
                      & BLIP2 & 1.07 ± 0.22 & 5.88 ± 0.24 & 73.81 ± 0.02 & 69.16 ± 0.03 & 70.76 ± 0.02 & 77.48 ± 0.05 &  \\
                      & BLIP2-FT & 2.41 ± 0.06 & 6.40 ± 0.26 & 80.10 ± 0.03 & 75.08 ± 0.15 & 77.03 ± 0.10 & 83.72 ± 0.12 &      \\ \midrule
&&&&&&\textbf{Non-Hispanic} & \textbf{Hispanic} \\
\multirow{4}{*}{\textbf{Ethnicity}} & CLIP & 15.83 ± 0.42 & 14.73 ± 0.54 & 77.27 ± 0.03 & 71.70 ± 0.08 & 77.51 ± 0.03 & 69.73 ± 0.13 & \\

                      & CLIP-FT       & 14.50 ± 0.72	& 22.49 ± 1.44 & 80.27 ± 0.08 &	76.31 ± 0.36 & 80.48 ± 0.07 & 75.30 ± 0.46 &  \\
                      & BLIP2   & 8.78 ± 1.35 & 16.56 ± 1.89 & 73.81 ± 0.02 &	68.75 ± 0.08 & 74.10 ± 0.02 & 66.74 ± 0.14 &  \\
                      & BLIP2-FT  & 16.64 ± 1.17 & 18.41 ± 1.98 & 80.10 ± 0.03 &	77.13 ± 0.09 & 80.25 ± 0.03 & 76.39 ± 0.12 &      \\ \midrule
&&&&&&\textbf{English} & \textbf{Spanish} & \textbf{Others} \\
\multirow{4}{*}{\textbf{Language}} & CLIP       & 13.57 ± 1.35 & 33.11 ± 0.53 & 77.27 ± 0.03 & 70.89 ± 0.21 & 77.25 ± 0.03 & 84.00 ± 0.16 & 75.02 ± 0.28 \\
                      & CLIP-FT       & 16.75 ± 1.08 & 15.74 ± 0.28 & 80.27 ± 0.08 & 67.06 ± 0.46 & 80.77 ± 0.07 & 74.43 ± 0.99 & 66.91 ± 0.17 \\
                      & BLIP2   & 22.40 ± 0.33 & 15.41 ± 1.41 & 73.81 ± 0.02 & 70.34 ± 0.64 & 73.40 ± 0.02 & 75.95 ± 0.87 & 76.19 ± 0.12 \\
                      & BLIP2-FT  & 14.08 ± 3.56 & 37.68 ± 2.51 & 80.10 ± 0.03 & 69.98 ± 0.56 & 80.62 ± 0.04 & 83.14 ± 1.18 & 69.51 ± 0.32 \\ 
\bottomrule
\end{tabular}}
\end{table*}

\noindent\textbf{Metrics:} To comprehensively understand the balance between model performance and fairness, we use multiple metrics for evaluation, including Demographic Parity Difference (DPD) \cite{Agarwal_ICML_2018,Agarwal_ICML_2019}, Difference in Equalized Odds (DEOdds) \cite{Agarwal_ICML_2018}, Area Under the Receiver Operating Characteristic Curve (AUC), Equity-Scaled AUC \cite{Luo_arXiv_2023}, and Group-wise AUC.
More details can be found in the supplementary material.


\subsection{VL Fairness Analysis}
\label{sec:vl_fairness_benchmark}

In this section, we present a comprehensive fairness analysis of two widely-used VL models, benchmarked across two different pre-training domains and four different protected attributes. Table~\ref{tab:vl_fairness_benchmark} presents the linear probing results, examining various performance (AUC) and fairness (DPD, DEOdds, ES-AUC) metrics, as well as reporting the group-wise AUC scores across the individual subgroups within each of the four protected attributes. We primarily focus on the ES-AUC metric in the subsequent analysis since it captures notions of both overall performance as well as fairness -- both of which are important for safety-critical medical applications (Sec.~\ref{sec:experimental_setup}). Next, we briefly discuss the disparities in VL performance across the various protected attributes, and the impact of different VL pre-training domains (natural vs. medical) and VL pre-training methods (CLIP vs. BLIP2) on model performance and fairness.

\noindent\textbf{Protected Attributes:} Across the four protected attributes -- race, gender, ethnicity, and language -- our results indicate that VL models exhibit the best performance-fairness trade-off on ethnicity and the worst on language with average ES-AUC scores (across the four VL models) of 73.47 and 69.57, respectively. A granular analysis reveals that in terms of racial subgroups, Asian patients consistently have the highest diagnostic performance, whereas Black patients have the lowest. Across genders, male patients are consistently better diagnosed than female patients. Moreover, non-Hispanic is the highest-performing subgroup across ethnicities, whereas Spanish speakers have the best diagnostic performance across the language subgroups. Some of these performance disparities could be attributed to the imbalances in the pre-training datasets. For instance, non-Hispanic patients make up 90.6\% of our dataset, potentially leading to improved performance in this subgroup. However, this is unlikely to be the only factor responsible for these performance disparities since the Asian, Male, and Spanish subgroups exhibit superior performances despite being the minority subgroups. This indicates that the pre-training of these models could potentially play a role in the biases exhibited by these models. We delve deeper into this in the subsequent analysis.

\begin{table*}[ht]
\centering
\scriptsize
\caption{Zero-shot transfer results of CLIP vs. FairCLIP, reporting the mean and standard deviation across three random seeds.
}
\vspace{-2ex}
\label{tab:zero_shot_clip}
\adjustbox{width=.9\textwidth}{
\begin{tabular}{llccccccc}
\toprule
\textbf{Attribute} & \textbf{Model}  & \textbf{DPD $\downarrow$} & \textbf{DEOdds $\downarrow$} & \textbf{AUC $\uparrow$} & \textbf{ES-AUC $\uparrow$} & \multicolumn{3}{c}{\textbf{Group-wise AUC $\uparrow$}} \\ \midrule
&&&&&&\textbf{Asian} & \textbf{Black} & \textbf{White} \\
\multirow{4}{*}{\textbf{Race}} & CLIP (ViT-B/16)    & 15.35 $\pm$ 6.50 & 15.11 $\pm$ 5.01 & 67.84 $\pm$ 0.90 & 61.67 $\pm$ 0.63 & 73.11 $\pm$ 2.74 & 70.78 $\pm$ 1.84 & 66.02 $\pm$ 0.60 \\
                       & FairCLIP (ViT-B/16)   & \textbf{6.07} $\pm$ 2.44 & \textbf{10.50} $\pm$ 2.73 & \textbf{70.24} $\pm$ 1.26 & 65.50 $\pm$ 2.60 & \textbf{74.83} $\pm$ 0.46 & 71.39 $\pm$ 0.66 & \textbf{69.17} $\pm$ 2.10 \\
                       & CLIP (ViT-L/14)    & 10.10 $\pm$ 9.44 & 10.79 $\pm$ 10.41 & 67.83 $\pm$ 2.92 & 63.53 $\pm$ 1.83 & 70.65 $\pm$ 4.58 & 70.12 $\pm$ 3.39 & 66.22 $\pm$ 2.97 \\
                       & FairCLIP (ViT-L/14)   & 17.79 $\pm$ 4.86 & 18.30 $\pm$ 2.07 & 69.88 $\pm$ 2.00 & \textbf{66.54} $\pm$ 1.73 & 71.78 $\pm$ 2.18 & \textbf{71.79} $\pm$ 2.13 & 68.67 $\pm$ 1.99  \\ 
                       \midrule
&&&&&&\textbf{Female} & \textbf{Male} \\
\multirow{4}{*}{\textbf{Gender}} & CLIP (ViT-B/16)    & 4.34 $\pm$ 0.66 & 9.95 $\pm$ 0.64 & 67.84 $\pm$ 0.90 & 63.21 $\pm$ 0.83 & 64.62 $\pm$ 0.83 & 71.96 $\pm$ 1.22
 \\
                       & FairCLIP (ViT-B/16)   & \textbf{0.84} $\pm$ 0.25 & \textbf{2.97} $\pm$ 2.07 & \textbf{69.76} $\pm$ 2.49 & 65.39 $\pm$ 2.39 & 66.81 $\pm$ 2.50 & \textbf{73.50} $\pm$ 2.41 \\
                       & CLIP (ViT-L/14)    & 2.93 $\pm$ 3.17 & 4.29 $\pm$ 4.05 & 67.83 $\pm$ 2.92 & 63.86 $\pm$ 2.36 & 65.13 $\pm$ 2.60 & 71.31 $\pm$ 3.24 \\
                       & FairCLIP (ViT-L/14)   & 5.82 $\pm$ 1.22 & 8.14 $\pm$ 2.62 & 69.74 $\pm$ 0.95 & \textbf{66.00} $\pm$ 1.55 & \textbf{67.29} $\pm$ 1.38 & 72.99 $\pm$ 0.83  \\  \midrule
&&&&&&\textbf{Non-Hispanic} & \textbf{Hispanic} \\
\multirow{4}{*}{\textbf{Ethnicity}} & CLIP (ViT-B/16)    & 8.86 $\pm$ 5.95 & 15.33 $\pm$ 5.18 & 67.84 $\pm$ 0.90 & 63.09 $\pm$ 0.26 & 68.08 $\pm$ 0.97 & 60.56 $\pm$ 0.30 \\
                       & FairCLIP (ViT-B/16)   & 9.12 $\pm$ 3.25 & 14.30 $\pm$ 7.20 & 69.30 $\pm$ 1.87 & 63.33 $\pm$ 0.25 & 69.62 $\pm$ 2.02 & 60.19 $\pm$ 1.18 \\
                       & CLIP (ViT-L/14)    & \textbf{7.78} $\pm$ 5.50 & \textbf{11.56} $\pm$ 9.03 & 67.83 $\pm$ 2.92 & 62.20 $\pm$ 0.82 & 68.14 $\pm$ 3.07 & 59.13 $\pm$ 0.92
 \\
                       & FairCLIP (ViT-L/14)   & 9.44 $\pm$ 3.28 & 13.03 $\pm$ 5.56 & \textbf{69.87} $\pm$ 1.05 & \textbf{64.90} $\pm$ 2.05 & \textbf{70.16} $\pm$ 1.05 & \textbf{62.43} $\pm$ 2.99  \\  \midrule
&&&&&&\textbf{English} & \textbf{Spanish} & \textbf{Others} \\
\multirow{4}{*}{\textbf{Language}} & CLIP (ViT-B/16)    & 11.11 $\pm$ 2.42 & 16.97 $\pm$ 2.73 & 67.84 $\pm$ 0.90 & 60.19 $\pm$ 3.47 & 67.88 $\pm$ 1.15 & 61.93 $\pm$ 4.57 & 61.89 $\pm$ 2.90 \\
                       & FairCLIP (ViT-B/16)   & \textbf{7.34} $\pm$ 4.63 & 17.15 $\pm$ 11.13 & \textbf{70.08} $\pm$ 1.14 & \textbf{62.31} $\pm$ 0.96 & \textbf{70.22} $\pm$ 1.27 & \textbf{68.47} $\pm$ 5.49 & \textbf{63.47} $\pm$ 2.13 \\
                       & CLIP (ViT-L/14)    & 7.62 $\pm$ 5.39 & \textbf{10.84} $\pm$ 9.84 & 67.83 $\pm$ 2.92 & 61.29 $\pm$ 1.28 & 67.77 $\pm$ 3.18 & 63.83 $\pm$ 3.20 & 62.24 $\pm$ 2.25 \\
                       & FairCLIP (ViT-L/14)   & 11.63 $\pm$ 3.53 & 11.05 $\pm$ 5.66 & 68.68 $\pm$ 2.26 & 61.50 $\pm$ 2.29 & 68.71 $\pm$ 2.31 & 65.06 $\pm$ 3.96 & 61.18 $\pm$ 1.20 \\ 
\bottomrule
\end{tabular}}
\end{table*}
\noindent\textbf{Natural vs. Medical Pre-training:} Here, we compare the VL models pre-trained on two different domains -- natural vs. medical VL datasets. Comparing the two pre-training domains, we observe that pre-training on medical data (i.e., paired fundus images + clinical notes) improves the performance-fairness trade-off across all protected attributes except language. Concretely, we see the most benefit on ethnicity, with improvements of as much as 8.38 in terms of ES-AUC. A granular analysis of the individual subgroups shows that the White and Hispanic subgroups witness the most improvement from medical pre-training, with improvements of as much as 6.75 and 9.65 ES-AUC in the former and latter, respectively. This indicates that medical VL pre-training could be an effective strategy for improving the performances of under-performing subgroups, thereby effectively reducing the performance disparities across different protected attributes.

\noindent\textbf{CLIP vs. BLIP2:} Next, we investigate the impact of different VL pre-training methods on the performance disparities of the downstream models. In order to ensure a fair comparison, we use the same ViT-L/14 architecture for the vision encoders of both models. Unlike CLIP, BLIP2 is a parameter-efficient VL method that keeps the vision encoder frozen and leverages a Q-former to learn the most suitable image representations corresponding to the text. From the natural pre-training results in Table~\ref{tab:vl_fairness_benchmark}, we note that CLIP consistently outperforms BLIP2 in terms of not only overall performance (77.27 vs. 73.81 AUC) but also fairness, as evidenced by the superior ES-AUC scores across all four protected attributes. However, we observe that medical pre-training largely closes this gap, with BLIP2 yielding similar results as CLIP (80.10 vs. 80.27 AUC). This large boost in BLIP2 performance indicates that (1) the parameter-efficient training in BLIP2 is effective for adapting to the medical VL data, and that (2) the clinical notes are useful for guiding BLIP2 to extract meaningful visual features from the frozen vision encoder. Among the medical pre-trained CLIP and BLIP2 models, CLIP exhibits a better performance-fairness trade-off than BLIP2 on the race and gender attributes, whereas BLIP2 yields more favorable results on the ethnicity and language attributes.

In summary, we observe that all VL models exhibit biases, with Asian, Male, Non-Hispanic, and Spanish being the preferred subgroups across the protected attributes of race, gender, ethnicity, and language, respectively. Medical pre-training enhances the performance-fairness trade-off across all attributes except language. Furthermore, different VL pre-training methods exhibit varying strengths, with CLIP outperforming on race and gender, whereas BLIP2 yields superior results on ethnicity and language.

\subsection{CLIP vs. FairCLIP}
\label{sec:fairclip}
Table~\ref{tab:zero_shot_clip} compares the zero-shot performance of CLIP against FairCLIP across two different architectures (ViT-B/16 and ViT-L/14) and four different protected attributes. Both CLIP and FairCLIP are fine-tuned with the pairs of images and clinical notes without supervised information (\ie, labels). Then, the resulting models are evaluated in the classification task.
CLIP exhibits notable disparities in group-wise AUC for attributes such as race, gender, ethnicity, and language, indicating the presence of bias in glaucoma detection. Regarding the racial groups, FairCLIP (ViT-B/16) gains a significant improvement in fairness, reducing DPD and DEOdds to 6.07 and 10.50, respectively, while increasing the AUC for the White group to 69.17. In terms of gender disparity, FairCLIP (ViT-B/16) addresses the bias effectively, elevating the female AUC to 66.81 from 64.62, indicating a substantial enhancement in gender. Similar enhancements by FairCLIP are observed in ethnicity and language groups.





Overall, FairCLIP demonstrates a significant improvement over CLIP in terms of both fairness metrics (DPD, DEOdds) as well as ES-AUC and AUC scores across various demographic subgroups. The supplementary material shows more end-to-end fine-tuning results, further validating the effectiveness of FairCLIP. These empirical findings suggest that optimizing the distance between the overall sample distribution and the distribution w.r.t. specific subgroups effectively improves fairness, indicating a promising direction in addressing and mitigating inherent biases.

\subsection{Ablation Studies}
\label{sec:ablations}

\begin{figure}[t]
\centering
    \begin{subfigure}[b]{0.15\textwidth}
            \centering
            \includegraphics[width=1\linewidth]
    {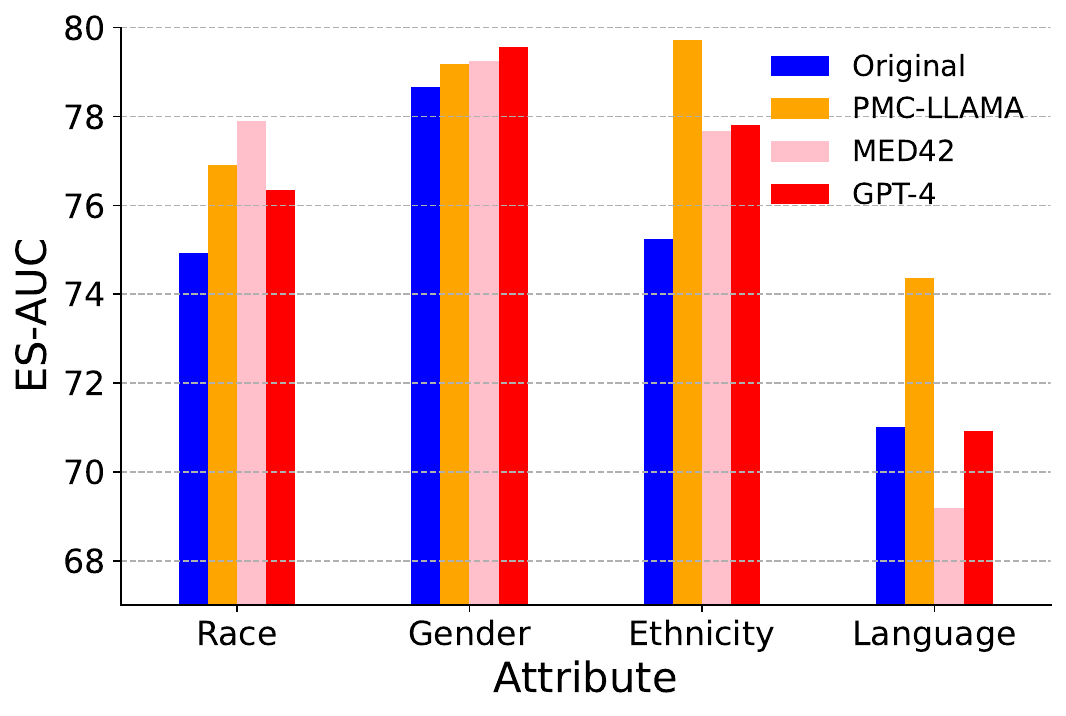}
    \caption{}
    \label{fig:summarization_methods}
    \end{subfigure} 
    \begin{subfigure}[b]{0.15\textwidth}
            \centering
            \includegraphics[width=1\textwidth]{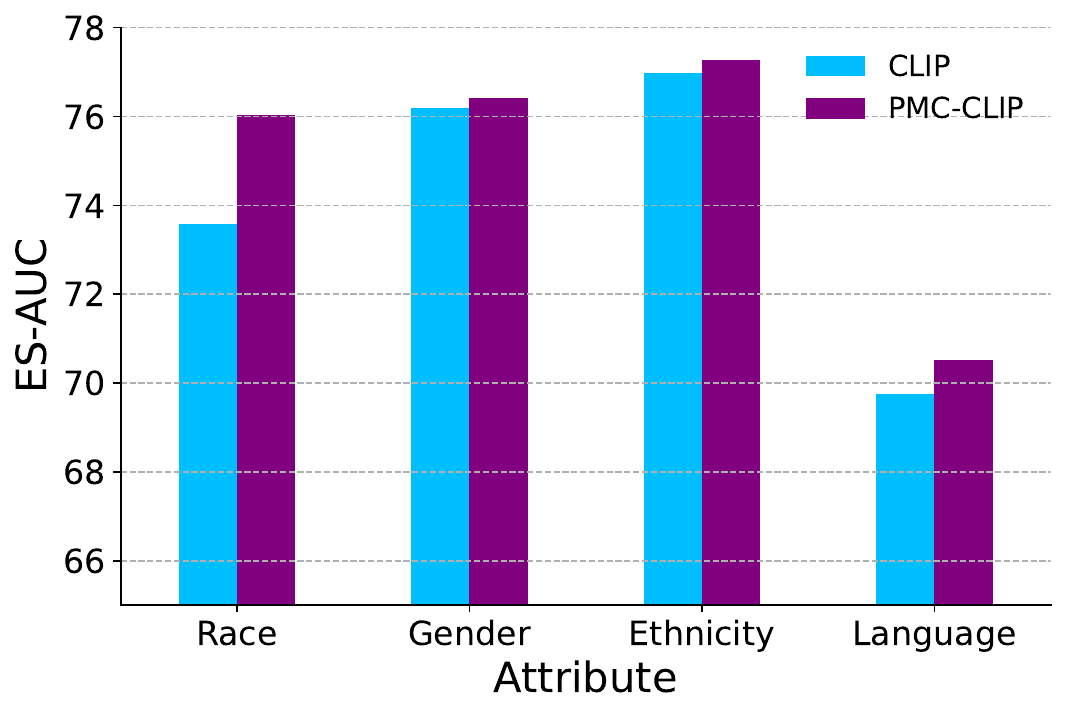}
        \caption{}
    \label{fig:ablation_pmcclip}
    \end{subfigure} 
    \begin{subfigure}[b]{0.15\textwidth}
            \centering
            \includegraphics[width=1\textwidth]{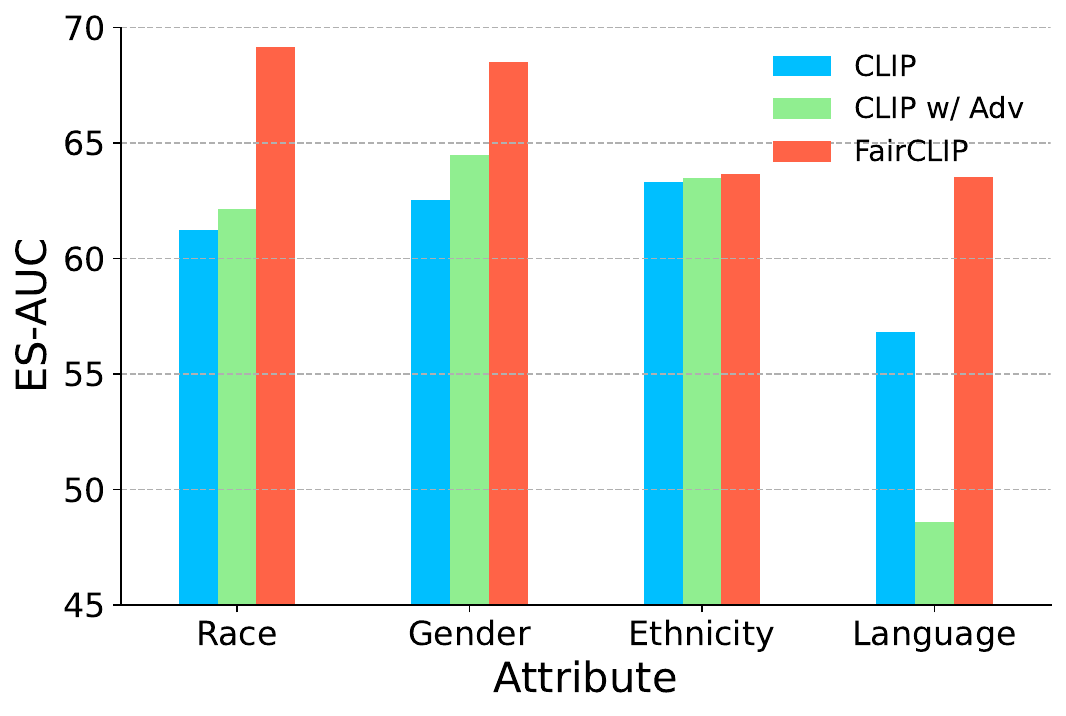}
        \caption{}
    \label{fig:ablation_advloss}
    \end{subfigure} 
    \caption{Extensive analyses on Harvard-FairVLMed, including (a) Effects of various LLM summarizations on the performance-fairness trade-off of BLIP2, (b) Effects of using pre-trained vision encoders from natural (CLIP) and medical (PMC-CLIP) domains on the performance-fairness trade-off of BLIP2, and (c) Performance comparison among CLIP, CLIP with adversarial loss (CLIP w/ Adv), and FairCLIP.}
    \vspace{-4ex}
\end{figure}

\noindent\textbf{Clinical Note Summarization:} VL models are usually trained with images and captions, which tend to be quite short. In contrast, the clinical notes in our Harvard-FairVLMed dataset are fairly lengthy, capturing a lot more nuanced information. Hence, we summarize these clinical notes in order to align our setup with the standard VL frameworks while retaining essential medical information. We use three different types of LLMs for summarization. In addition to the SOTA GPT-4~\cite{openai2023gpt4} model, we also use two LLMs specialized for the medical domain -- PMC-LLAMA~\cite{wu2023pmc} and MED42\footnote{https://huggingface.co/m42-health/med42-70b}. We use the following prompt for summarization for all three LLMs: \textit{Summarize the key details, including the presence of glaucoma, from the clinical note within 180 characters.} Figure~\ref{fig:summarization_methods} compares the performance-fairness trade-off of BLIP2 models trained via the three summarized notes against the original notes. Across all protected attributes except language, we note that all three summarization methods yield a consistent improvement in ES-AUC. Particularly, MED42 yields the best performance across race, GPT-4 yields the best performance across gender, whereas PMC-LLAMA yields the best performance across both ethnicity and language. Overall, the medical pre-trained LLMs (PMC-LLAMA and MED42) yield superior results, highlighting the efficacy of domain-specific LLMs for clinical note summarization.

\noindent\textbf{Vision vs. Multimodal Features:} In order to decouple the benefits of image and text features, we conduct linear probing on the BLIP2 pre-trained models using either vision-only or (vision + language) features. Table~\ref{tab:vision_vs_multimodal_features} presents the performance-fairness trade-off in terms of ES-AUC. We note that the multimodal features consistently improve the performance-fairness trade-off across all protected attributes except language. This highlights that the VL models make effective use of the clinical textual features, with the most appreciable gains observed on the race attribute.

\noindent\textbf{Natural vs. Medical Vision Encoder:} To investigate the impact of different vision encoders on model fairness in BLIP2, we utilize two different pre-trained encoders -- 1) CLIP trained on the natural domain, whereas 2) PMC-CLIP trained on the medical domain. The results in Figure~\ref{fig:ablation_pmcclip} reveal that PMC-CLIP outperforms CLIP across all four protected attributes, with the most appreciable gains on the racial subgroups. We note that medical-specific LLM summarizers and vision encoders consistently improve the performance-fairness trade-off of VL models, with the most appreciable improvements across the race attribute.

\noindent\textbf{Comparison with Adversary Fairness:} Beutel et al.~\cite{beutel2017data} introduce a fairness approach that employs adversarial loss to prevent the model from inaccurately predicting sensitive attributes. This method aims to ensure that the model predicts the label of an image without relying on its sensitive attributes, thereby reducing bias in classification. Figure~\ref{fig:ablation_advloss} shows the performance comparison among CLIP, CLIP with adversarial loss (CLIP w/ Adv), and FairCLIP.
The performance of CLIP with adversarial training (CLIP w/ Adv) does not consistently surpass that of the standard CLIP across all attributes. In contrast, FairCLIP consistently outperforms CLIP. This variation in performance can be attributed to the inherent challenges of adversarial training in maintaining equivalent prediction accuracy for each attribute. On the other hand, FairCLIP utilizes Sinkhorn loss, which effectively encourages uniformity in the distribution of all samples relative to the distributions corresponding to each group.

\begin{table}[t]
\centering
\scriptsize
\caption{Impact of vision-only and (vision + language) features on the performance-fairness trade-off of linear probing via BLIP2.}
\vspace{-2ex}
\label{tab:vision_vs_multimodal_features}
\resizebox{.9\columnwidth}{!}{%
\begin{tabular}{lllllll}
\toprule
\textbf{Attribute} & \textbf{V} & \textbf{L} & \textbf{ES-AUC $\uparrow$} & \multicolumn{3}{c}{\textbf{Group-wise AUC $\uparrow$}} \\ \midrule
&&&&\textit{Asian} & \textit{Black} & \textit{White} \\
\multirow{2}{*}{\textbf{Race}} & \cmark & \xmark & 73.76 & 82.08 & 74.35 & 81.03 \\
                      & \cmark  & \cmark & 76.70 & 79.79 & 77.77 & 82.37 \\ \midrule
&&&&\textit{Female} & \textit{Male} \\
\multirow{2}{*}{\textbf{Gender}} & \cmark & \xmark & 75.22 & 77.13 & 83.66 \\
                      & \cmark  & \cmark & 79.15 & 80.21 & 83.38 \\ \midrule
&&&&\textit{Non-Hispanic} & \textit{Hispanic} \\
\multirow{2}{*}{\textbf{Ethnicity}} & \cmark & \xmark & 77.18 & 80.28 & 76.46 \\
                      & \cmark  & \cmark & 75.54 & 81.95 & 73.85 \\ \midrule
&&&&\textit{English} & \textit{Spanish} & \textit{Others} \\
\multirow{2}{*}{\textbf{Language}} & \cmark & \xmark & 69.88 & 80.64 & 83.52 & 69.36 \\
                      & \cmark & \cmark & 66.35 & 82.32 & 69.89 & 71.01 \\ 
\bottomrule
\end{tabular}
}
\end{table}

\section{Conclusion}
Given the critical need for fairness in healthcare-focused settings, we introduce the first fair vision-language medical dataset (\textit{Harvard-FairVLMed}) for studying the fairness of medical VL foundation models. Our comprehensive fairness analysis on Harvard-FairVLMed reveals significant biases in all VL models. To address these biases, we propose FairCLIP, an optimal transport-based approach that effectively balances both performance and fairness. To aid future research in this area, we make our dataset and code publicly available at \url{https://ophai.hms.harvard.edu/datasets/harvard-fairvlmed10k}.


\section*{Acknowledgement}
This work was supported by NIH R00 EY028631, NIH R21 EY035298, NIH P30 EY003790, Research To Prevent Blindness International Research Collaborators Award, and Alcon Young Investigator Grant. We also acknowledge the generous funding resources provided by NYU Abu Dhabi with code AD131.

{
    \small
    \bibliographystyle{ieeenat_fullname}
    \bibliography{sec/ref}
}

\setcounter{page}{1}
\maketitlesupplementary

\section{Experimental Setup}


\subsection{Pre-Training}
\label{sec:pretraining_setup}
We use the widely-adopted VL methods -- CLIP~\cite{Radford_ICML_2021} and BLIP2~\cite{li2023blip} -- for our analysis. For the natural pre-trained variants, we use the official checkpoints provided by CLIP and BLIP2. For the medical pre-trained variants, we pre-train both methods on our Harvard-FairVLMed dataset after initializing from the official checkpoints.
We fine-tune CLIP for 10 epochs using the Adam~\cite{kingma_ICLR_2014} optimizer, with a learning rate (lr) of 1e-5. The hyperparameters $\beta_1$ and $\beta_2$ are configured at 0.1, along with a weight decay of 6e-5 and batch size of 32. These specific hyper-parameters were selected after extensive tuning to achieve optimal performance with CLIP. FairCLIP uses the same aforementioned hyper-parameters, leveraging a batch size $|\mathcal{B}_{a}|$ of 32 to draw samples from each group. The Sinkhorn loss is integrated with CLIP's original loss function, using a weight of 1e-7. For BLIP2, we primarily focus on the vision-language representation learning stage (i.e., Stage 1) and use the official ViT-L/14 model from CLIP with FP32 precision as the frozen vision encoder. Following the official implementation, we use AdamW~\cite{Loshchilov_ICLR_2019} as the optimizer, with $\beta_1$, $\beta_2$, and weight decay set to 0.9, 0.98, and 0.05, respectively. We also use a cosine lr decay with a max, min, and warmup lr of 1e-4, 1e-5, and 1e-6, respectively. Moreover, we apply random resized cropping (224$\times$224) and random horizontal flipping to the fundus images, whereas we utilize the BLIP caption augmentations to pre-process the clinical notes, with maximum words set to 50. Finally, all models are pre-trained on the paired fundus images and clinical notes from Harvard-FairVLMed using a batch size of 32 for 50 epochs on a single V100 GPU. These pre-trained CLIP and BLIP2 models are then used for the subsequent linear probing and zero-shot evaluation.

\subsection{Metrics}

To comprehensively understand the balance between model performance and fairness, we use multiple metrics for evaluation, including Demographic Parity Difference (DPD)~\cite{Agarwal_ICML_2018,Agarwal_ICML_2019}, Difference in Equalized Odds (DEOdds)~\cite{Agarwal_ICML_2018}, Area Under the Receiver Operating Characteristic Curve (AUC), Equity-Scaled AUC~\cite{Luo_arXiv_2023}, and Group-wise AUC. Particularly, DPD and DEOdds are widely used fairness metrics that focus on the fairness of the model's predictions, ensuring that no group is systematically advantaged or disadvantaged. In contrast, AUC is a mainstream performance metric used in medical scenarios. Group-wise AUC is an intuitive and straightforward metric to reveal the discrepancy between groups.
In safety-critical medical applications, neither fairness nor performance alone is sufficient as the sole measurement criterion. Hence, ES-AUC is an effective metric that efficiently balances both performance and fairness. It offers a holistic evaluation, facilitating the analysis of the trade-off between these two essential criteria. ES-AUC is defined as:
\begin{equation*}
    \begin{split}
    \text{ES-AUC} = \frac{\text{AUC}}{1+\sum_{a}^{\mathcal{A}}| \text{AUC} - \text{AUC}_{a} |}
\end{split}
\label{eqn:es_metric}
\end{equation*}
where $\mathcal{A}$ can be $\{\text{Asian},\text{Black},\text{White}\}$, $\{\text{Female},\text{Male}\}$, $\{\text{Non-Hispanic},\text{Hispanic}\}$, or $\{\text{English},\text{Spanish},\text{Others}\}$. A higher ES-AUC score indicates that the model achieves not only greater performance but also simultaneously improves model equity.

\section{Results}
\subsection{CLIP vs. FairCLIP}
In addition to the zero-shot comparison of CLIP and FairCLIP presented in Section 5.3, Table~\ref{tab:supervised_clip} demonstrates their end-to-end fine-tuning results, further validating the effectiveness of FairCLIP. Once again, the performance is evaluated using the metrics DPD, DEOdds, AUC, ES-AUC, and group-wise AUC.

\begin{table*}[t]
\centering
\scriptsize
\caption{End-to-end fine-tuning results of CLIP vs. FairCLIP, reporting the mean and standard deviation across three random seeds.}
\label{tab:supervised_clip}
\adjustbox{width=.8\textwidth}{
\begin{tabular}{llccccccc}
\toprule
\textbf{Attribute} & \textbf{Model}  & \textbf{DPD $\downarrow$} & \textbf{DEOdds $\downarrow$} & \textbf{AUC $\uparrow$} & \textbf{ES-AUC $\uparrow$} & \multicolumn{3}{c}{\textbf{Group-wise AUC $\uparrow$}} \\ \midrule
&&&&&&\textbf{Asian} & \textbf{Black} & \textbf{White} \\
\multirow{4}{*}{\textbf{Race}} & CLIP (ViT-B/16)    & \textbf{5.85} $\pm$ 3.39 & 10.68 $\pm$ 3.75 & 81.19 $\pm$ 0.44 & 75.07 $\pm$ 1.36 & 84.82 $\pm$ 1.77 & 77.15 $\pm$ 1.17 & 81.73 $\pm$ 0.61 \\
                       & FairCLIP (ViT-B/16)   & 11.38 $\pm$ 4.23 & 10.53 $\pm$ 3.10 & 81.70 $\pm$ 0.34 & \textbf{76.85} $\pm$ 0.64 & 83.30 $\pm$ 1.09 & 77.35 $\pm$ 0.87 & 82.07 $\pm$ 0.28 \\
                       & CLIP (ViT-L/14)    & 7.39 $\pm$ 1.98 & 10.59 $\pm$ 1.64 & 80.21 $\pm$ 1.43 & 75.37 $\pm$ 1.03 & 82.04 $\pm$ 2.26 & 76.29 $\pm$ 1.73 & 80.89 $\pm$ 1.42 \\
                       & FairCLIP (ViT-L/14)   & 8.67 $\pm$ 4.32 & \textbf{8.84} $\pm$ 5.24 & \textbf{81.80} $\pm$ 0.19 & 76.70 $\pm$ 1.74 & \textbf{84.87} $\pm$ 1.05 & \textbf{78.52} $\pm$ 1.37 & \textbf{82.17} $\pm$ 0.41  \\ \midrule
&&&&&&\textbf{Female} & \textbf{Male} \\
\multirow{4}{*}{\textbf{Gender}} & CLIP (ViT-B/16)    & 1.89 $\pm$ 1.65 & 6.78 $\pm$ 2.88 & 81.19 $\pm$ 0.44 & 77.47 $\pm$ 0.51 & 78.96 $\pm$ 0.30 & 83.78 $\pm$ 0.96 \\
                       & FairCLIP (ViT-B/16)   & \textbf{1.72} $\pm$ 0.36 & \textbf{5.59} $\pm$ 0.12 & \textbf{81.88} $\pm$ 0.30 & \textbf{78.46} $\pm$ 0.31 & \textbf{79.84} $\pm$ 0.25 & \textbf{84.20} $\pm$ 0.33 \\
                       & CLIP (ViT-L/14)    &  1.85 $\pm$ 0.95 & 6.73 $\pm$ 1.39 & 80.21 $\pm$ 1.43 & 76.39 $\pm$ 1.60 & 77.92 $\pm$ 1.56 & 82.93 $\pm$ 1.25 \\
                       & FairCLIP (ViT-L/14)   & 2.26 $\pm$ 1.28 & 7.58 $\pm$ 2.59 & 81.07 $\pm$ 0.78 & 77.36 $\pm$ 0.27 & 78.86 $\pm$ 0.44 & 83.66 $\pm$ 1.27  \\  \midrule
&&&&&&\textbf{Non-Hispanic} & \textbf{Hispanic} \\
\multirow{4}{*}{\textbf{Ethnicity}} & CLIP (ViT-B/16)    & \textbf{9.57} $\pm$ 2.34 & \textbf{11.35} $\pm$ 5.03 & 81.19 $\pm$ 0.44 & 76.09 $\pm$ 1.44 & 81.43 $\pm$ 0.53 & 74.68 $\pm$ 2.07 \\
                       & FairCLIP (ViT-B/16)   & 12.80 $\pm$ 2.01 & 14.49 $\pm$ 3.15 & \textbf{81.47} $\pm$ 0.15 & 78.22 $\pm$ 1.44 & \textbf{81.61} $\pm$ 0.21 & 77.42 $\pm$ 1.86 \\
                       & CLIP (ViT-L/14)    &  12.15 $\pm$ 3.21 & 15.08 $\pm$ 3.18 & 80.21 $\pm$ 1.43 & 75.79 $\pm$ 1.45 & 80.45 $\pm$ 1.45 & 74.61 $\pm$ 1.59 \\
                       & FairCLIP (ViT-L/14)   & 10.47 $\pm$ 0.96 & 13.62 $\pm$ 2.15 & \textbf{81.47} $\pm$ 0.58 & \textbf{79.08} $\pm$ 1.16 & 81.57 $\pm$ 0.64 & \textbf{78.52} $\pm$ 1.54 \\  \midrule
&&&&&&\textbf{English} & \textbf{Spanish} & \textbf{Others} \\
\multirow{4}{*}{\textbf{Language}} & CLIP (ViT-B/16)    & 13.12 $\pm$ 3.49 & 22.10 $\pm$ 3.77 & 81.19 $\pm$ 0.44 & 70.12 $\pm$ 1.71 & 81.64 $\pm$ 0.43 & \textbf{80.59} $\pm$ 5.00 & 70.62 $\pm$ 3.25 \\
                       & FairCLIP (ViT-B/16)   & 15.29 $\pm$ 1.83 & \textbf{21.14} $\pm$ 4.88 & \textbf{81.71} $\pm$ 0.28 & 71.74 $\pm$ 1.26 & \textbf{82.21} $\pm$ 0.30 & 79.36 $\pm$ 1.89 & 70.63 $\pm$ 0.29 \\
                       & CLIP (ViT-L/14)    &  \textbf{10.95} $\pm$ 5.92 & 26.58 $\pm$ 9.41 & 80.21 $\pm$ 1.43 & 70.77 $\pm$ 1.64 & 80.61 $\pm$ 1.42 & 78.12 $\pm$ 3.96 & 71.00 $\pm$ 1.48 \\
                       & FairCLIP (ViT-L/14)   & 15.81 $\pm$ 4.49 & 25.18 $\pm$ 11.78 & 81.22 $\pm$ 0.42 & \textbf{74.44} $\pm$ 1.22 & 81.41 $\pm$ 0.36 & \textbf{80.59} $\pm$ 4.38 & \textbf{75.65} $\pm$ 0.88 \\ 
\bottomrule
\end{tabular}}
\end{table*}

In terms of the racial subgroups, both CLIP and FairCLIP show varied performance. For instance, in the ViT-B/16 setting, CLIP achieves a lower DPD (5.85) compared to FairCLIP (11.38), indicating a better balance in outcomes across races. However, FairCLIP (ViT-L/14) outperforms CLIP in AUC and group-wise AUC for the Asian and Black groups, suggesting a more equitable performance across these racial categories. For the gender subgroups, FairCLIP consistently outperforms CLIP in both DPD and DEOdds, indicating a more balanced performance between the male and female subgroups. The AUC scores are also higher for FairCLIP, with the ViT-B/16 achieving an AUC of 81.88 and a higher group-wise AUC for both genders. In terms of ethnicity, FairCLIP generally achieves higher AUC scores than CLIP. Notably, FairCLIP (ViT-L/14) shows a significant improvement in ES-AUC (79.08) and group-wise AUC for Hispanic groups. Lastly, for the language subgroups, FairCLIP shows a slightly better performance in terms of AUC and group-wise AUC for English and Spanish speakers. However, both models struggle with the ``Others'' language group, with FairCLIP (ViT-L/14) showing a notable improvement in ES-AUC (74.44).

Overall, similar to the results presented in Section 5.3, we observe that our proposed method FairCLIP consistently outperforms CLIP.




\subsection{Dataset Analysis}
\begin{figure}[!h]
    \centering
    \vspace{-3ex}
    \begin{subfigure}{.14\textwidth}
      \centering
      \includegraphics[width=\linewidth]{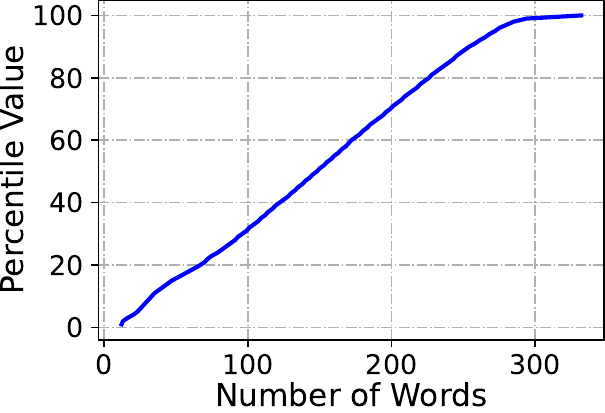}
      \caption{}
      \label{fig:HMJD_1}
    \end{subfigure} \hfill
    \begin{subfigure}{.12\textwidth}
      \centering
      \includegraphics[width=\linewidth]{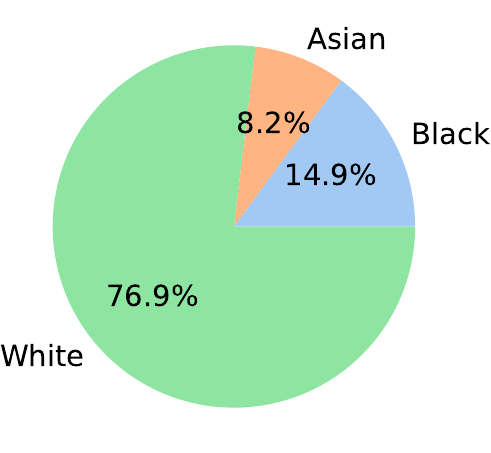}
      \caption{}
      \label{fig:HMJD_2}
    \end{subfigure} \hfill
    \begin{subfigure}{.11\textwidth}
      \centering
      \includegraphics[width=\linewidth]{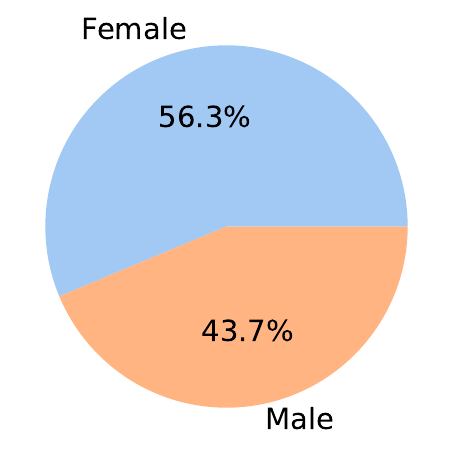}
      \caption{}
      \label{fig:HMJD_3}
    \end{subfigure}
    \vspace{-2ex}
    \caption{(a) Distribution of words in the clinical notes, (b) Prevalence of subjects across race, (c) Prevalence of subjects across gender.}
    \vspace{-3ex}
\end{figure}

To supplement the details for our Harvard-FairVLMed dataset presented in the main paper, here we provide additional analyses representing the distribution of words in the clinical notes (Figure~\ref{fig:HMJD_1}), and the prevalence of subjects across the race and gender attributes (Figure~\ref{fig:HMJD_2} and~\ref{fig:HMJD_3}), respectively.

\subsection{Ablation Studies}
\begin{figure}[h!]
    \centering
    \begin{subfigure}{.25\textwidth}
      \centering
      \includegraphics[width=\linewidth]{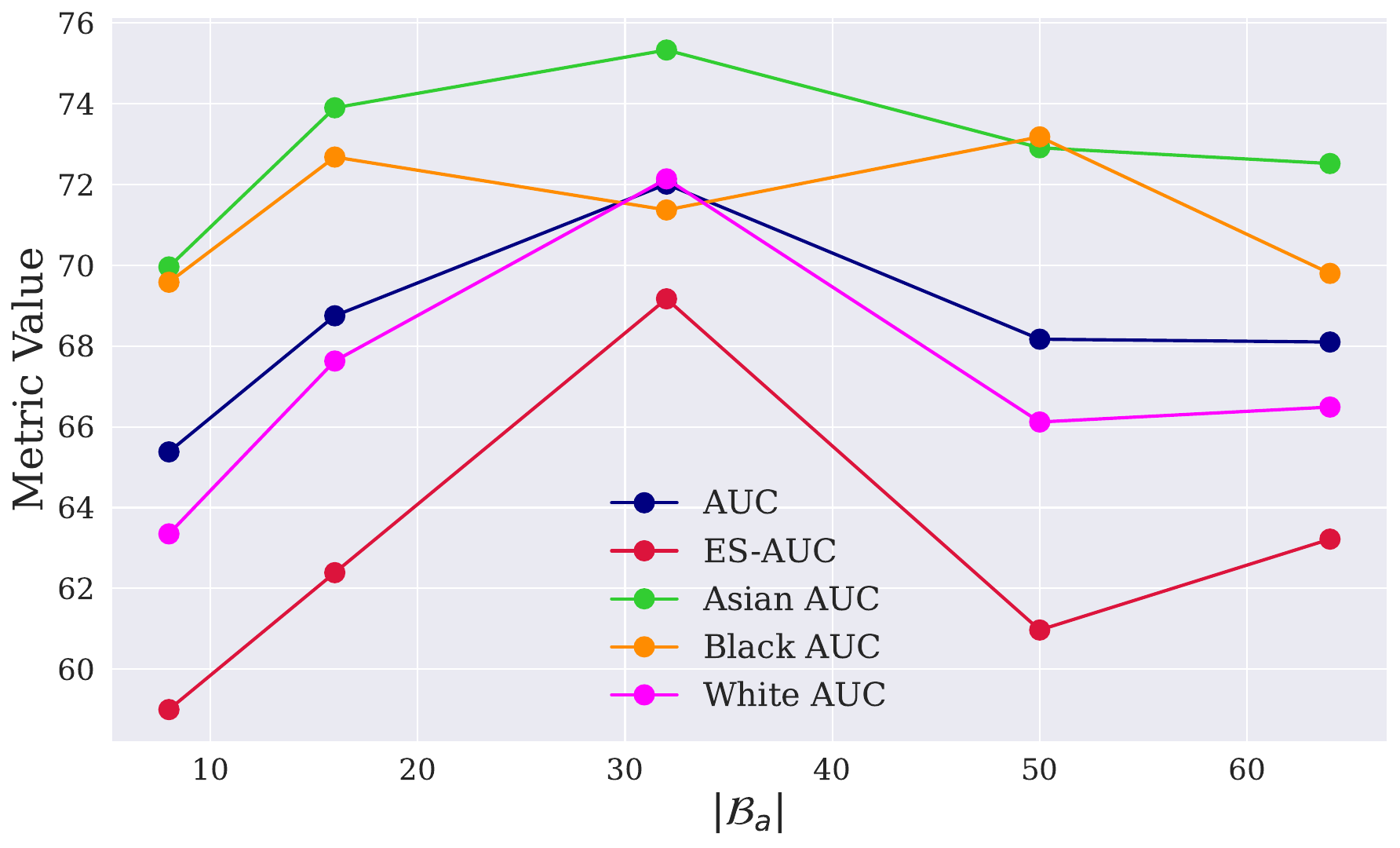}
        \caption{}
    \label{fig:ablation_bz}
    \end{subfigure} \hfill
    \begin{subfigure}{.215\textwidth}
      \centering
      \includegraphics[width=\linewidth]{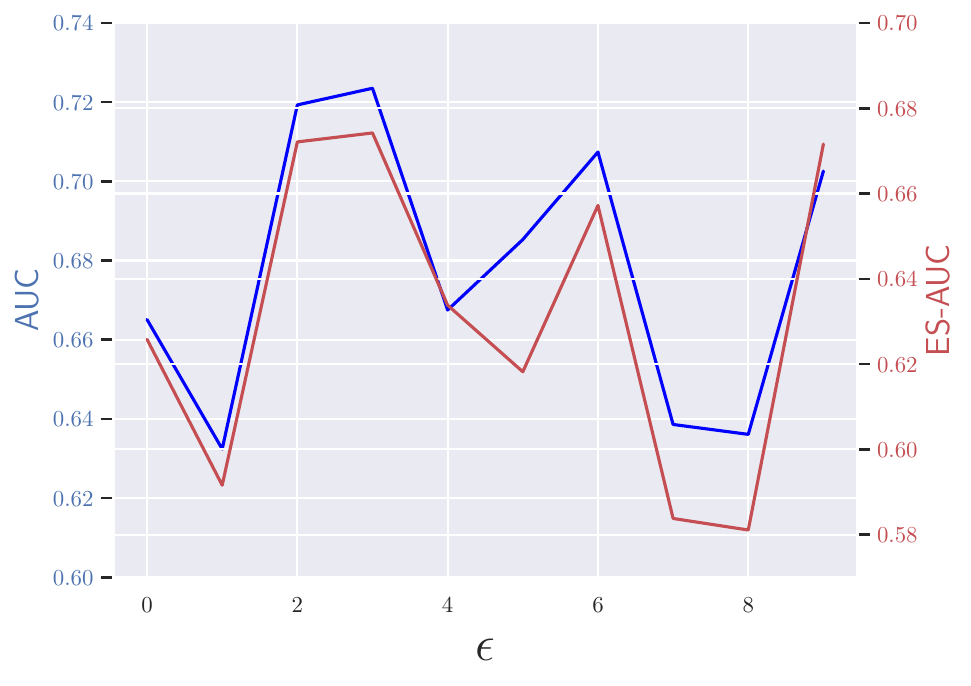}
      \caption{}
      \label{fig:HMJD_4}
    \end{subfigure} \\
    \begin{subfigure}{.22\textwidth}
      \centering
      \includegraphics[width=\linewidth]{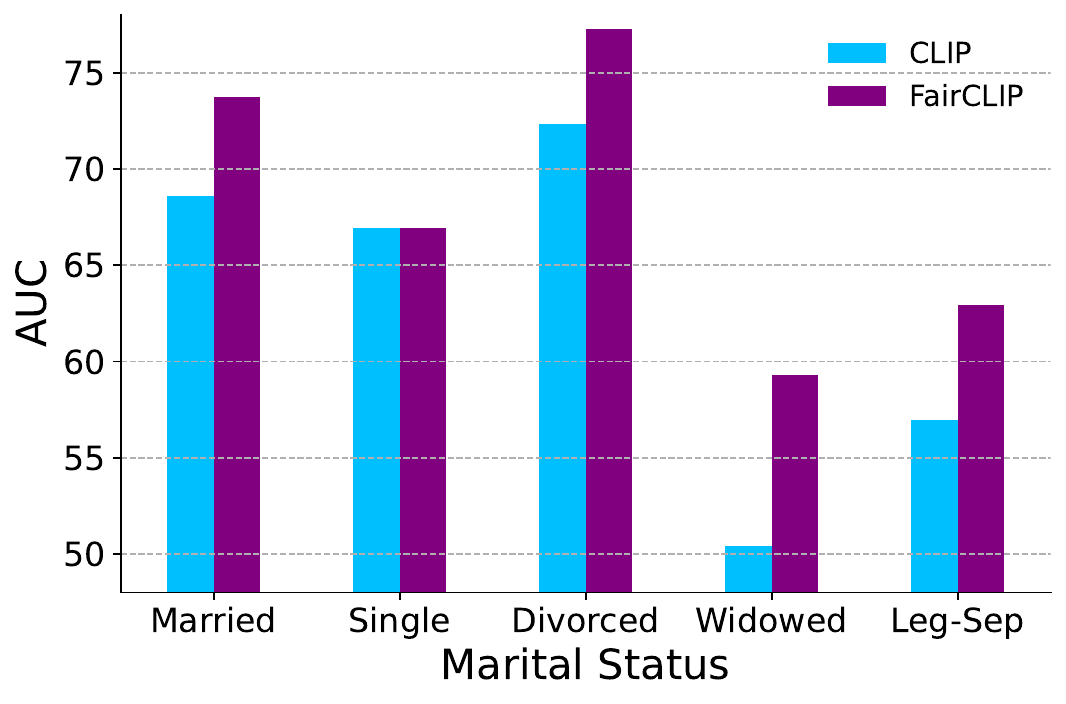}
      \caption{}
      \label{fig:kMiP_1}
    \end{subfigure} \hfill
    \begin{subfigure}{.22\textwidth}
      \centering
      \includegraphics[width=\linewidth]{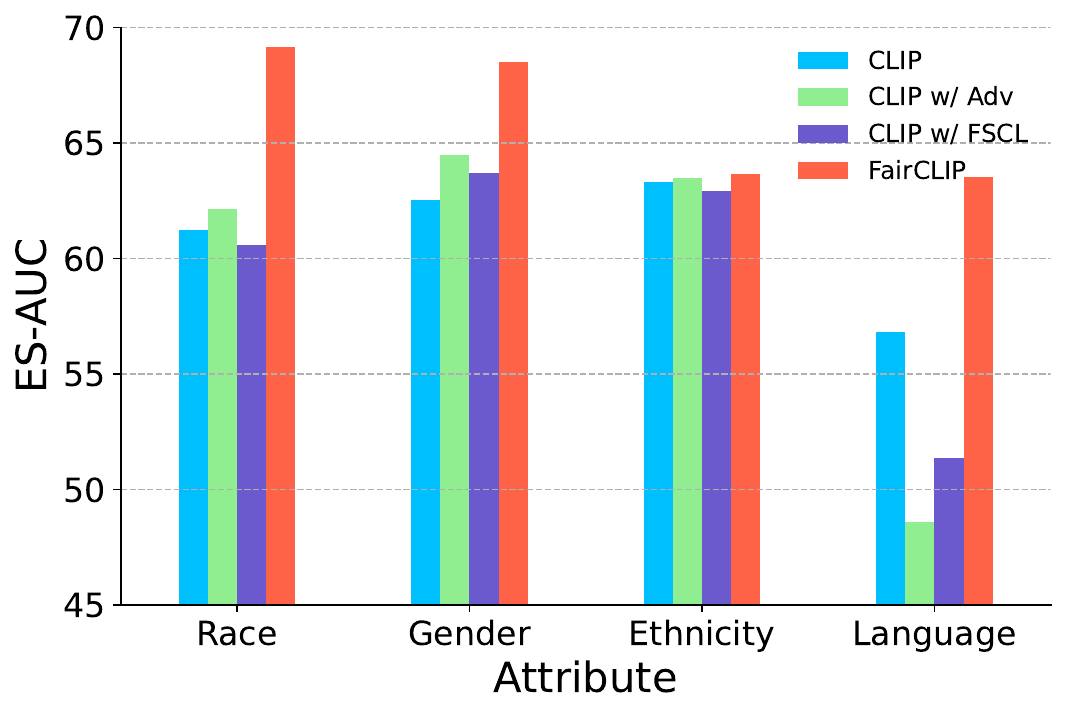}
      \caption{}
      \label{fig:kMiP_2}
    \end{subfigure}
    \vspace{-2ex}
    \caption{(a) Ablation study of using various $|\mathcal{B}_{a}|$ in FairCLIP, (b) Ablation study on the effects of $\epsilon$ on model performance, (c) Fairness results based on marital status, (d) Comparison of FairCLIP against other fairness algorithms.}
    \vspace{-3ex}
\end{figure}

In addition to the ablation studies presented in Section 5.4, we also study the effect of $|\mathcal{B}_{a}|$ in FairCLIP on model performance. From the results in Figure \ref{fig:ablation_bz}, we observe that $|\mathcal{B}_{a}| = 32$ achieves desired performance.

Moreover, we also present detailed results for the clinical note summarization, vision vs. multimodal features, and natural vs. medical vision encoder ablation studies in Tables~\ref{tab:summarized_notes},~\ref{tab:vision_vs_multimodal}, and~\ref{tab:natural_vs_medical_vision_encoders}, respectively. For a comprehensive discussion of these ablation studies, please refer to Section 5.4 in the main paper.

Furthermore, we present an ablation study on the effects of $\epsilon$ on model performance in Figure~\ref{fig:HMJD_4}. Also, we include additional fairness results based on marital status in Figure~\ref{fig:kMiP_1}. Lastly, we provide a comparison of FairCLIP against other fairness algorithms in Figure~\ref{fig:kMiP_2}.


\begin{table*}[ht]
\centering
\scriptsize
\caption{Impact of various LLM summarizations on the performance-fairness trade-off of BLIP2.}
\label{tab:summarized_notes}
\begin{tabular}{lllllllll}
\toprule
\textbf{Attribute} & \textbf{Clinical Notes}  & \textbf{DPD $\downarrow$} & \textbf{DEOdds $\downarrow$} & \textbf{AUC $\uparrow$} & \textbf{ES-AUC $\uparrow$} & \multicolumn{3}{c}{\textbf{Group-wise AUC $\uparrow$}} \\ \midrule
&&&&&&\textbf{Asian} & \textbf{Black} & \textbf{White} \\
\multirow{4}{*}{\textbf{Race}} & Original    & 8.36  &  11.28  &  80.13  &  73.76  &  82.08  & 74.35 & 81.03 \\
                       & PMC-LLAMA   & 4.38  &  12.71  &  80.11  &  72.63  &  83.77  & 74.20 & 80.84 \\
                       & MED42    & 6.26  &  14.49  &  80.36  &  73.59  &  82.93  & 74.60 & 81.23 \\
                       & GPT-4   & 5.30  &  6.24  &  79.34  &  74.39  &  82.45  & 76.15 & 79.70 \\ \midrule
&&&&&&\textbf{Female} & \textbf{Male} \\
\multirow{4}{*}{\textbf{Gender}} & Original    & 2.34  &  6.56  &  80.13  &  75.22  &  77.13  & 83.66 &  \\
                       & PMC-LLAMA   & 1.72  &  7.92  &  80.11  &  74.20  &  76.43  & 84.39 &  \\
                       & MED42    & 0.72  &  4.20  &  80.36  &  76.19  &  77.84  & 83.31 &  \\
                       & GPT-4   & 3.18  &  9.51  &  79.34  &  73.49  &  75.67  & 83.64 &  \\ \midrule
&&&&&&\textbf{Non-Hispanic} & \textbf{Hispanic} \\
\multirow{4}{*}{\textbf{Ethnicity}} & Original    & 16.26  &  20.59  &  80.13  &  77.18  &  80.28  & 76.46 &  \\
                       & PMC-LLAMA   & 14.96  &  16.17  &  80.11  &  76.24  &  80.24  & 75.17 &  \\
                       & MED42    & 16.32  &  18.25  &  80.36  &  76.98  &  80.49  & 76.11 &  \\
                       & GPT-4   & 16.55  &  15.83  &  79.34  &  76.40  &  79.44  & 75.59 &  \\ \midrule
&&&&&&\textbf{English} & \textbf{Spanish} & \textbf{Others} \\
\multirow{4}{*}{\textbf{Language}} & Original    & 11.47  &  39.13  &  80.13  &  69.88  &  80.64  & 83.52 & 69.36 \\
                       & PMC-LLAMA   & 9.15  &  34.78  &  80.11  &  71.71  &  80.60  & 78.13 & 70.88 \\
                       & MED42    & 21.78  &  22.28  &  80.36  &  69.76  &  80.70  & 72.16 & 73.71 \\
                       & GPT-4   & 14.65  &  39.13  &  79.34  &  69.52  &  79.89  & 77.27 & 67.84 \\ 
\bottomrule
\end{tabular}
\end{table*}

\begin{table*}[ht]
\centering
\scriptsize
\caption{Impact of vision-only and (vision + language) features on the performance-fairness trade-off of linear probing via BLIP2.}
\label{tab:vision_vs_multimodal}
\begin{tabular}{llllllllll}
\toprule
\textbf{Attribute} & \textbf{V} & \textbf{L} & \textbf{DPD $\downarrow$} & \textbf{DEOdds $\downarrow$} & \textbf{AUC $\uparrow$} & \textbf{ES-AUC $\uparrow$} & \multicolumn{3}{c}{\textbf{Group-wise AUC $\uparrow$}} \\ \midrule
&&&&&&&\textbf{Asian} & \textbf{Black} & \textbf{White} \\
\multirow{2}{*}{\textbf{Race}} & \cmark & \xmark & 6.26 & 14.49 & 80.36 & 73.59 & 82.93 & 74.60 & 81.23 \\
                               & \cmark & \cmark & 7.78 & 5.35 & 82.16 & 79.20 & 80.84 & 80.27 & 82.68 \\ \midrule
&&&&&&&\textbf{Female} & \textbf{Male} \\
\multirow{2}{*}{\textbf{Gender}} & \cmark & \xmark & 0.72 & 4.20 & 80.36 & 76.19 & 77.84 & 83.31 & \\
                                 & \cmark & \cmark & 1.12 & 3.87 & 82.16 & 79.56 & 80.66 & 83.93 & \\ \midrule
&&&&&&&\textbf{Non-Hispanic} & \textbf{Hispanic} \\
\multirow{2}{*}{\textbf{Ethnicity}} & \cmark & \xmark & 16.32 & 18.25 & 80.36 & 76.98 & 80.49 & 76.11 & \\
                                    & \cmark & \cmark & 16.19 & 15.69 & 82.16 & 78.98 & 82.29 & 78.26 & \\ \midrule
&&&&&&&\textbf{English} & \textbf{Spanish} & \textbf{Others} \\
\multirow{2}{*}{\textbf{Language}} & \cmark & \xmark & 21.78 & 22.28 & 80.36 & 69.76 & 80.70 & 72.16 & 73.71 \\
                                   & \cmark & \cmark & 15.08 & 21.73 & 82.16 & 66.27 & 82.76 & 68.18 & 72.77 \\ 
\bottomrule
\end{tabular}
\end{table*}

\begin{table*}[ht]
\centering
\scriptsize
\caption{Impact of using pre-trained vision encoders from natural (CLIP) and medical (PMC-CLIP) domains on the performance-fairness trade-off of BLIP2.}
\label{tab:natural_vs_medical_vision_encoders}
\begin{tabular}{lllllllll}
\toprule
\textbf{Attribute} & \textbf{Encoder Type}  & \textbf{DPD $\downarrow$} & \textbf{DEOdds $\downarrow$} & \textbf{AUC $\uparrow$} & \textbf{ES-AUC $\uparrow$} & \multicolumn{3}{c}{\textbf{Group-wise AUC $\uparrow$}} \\ \midrule
&&&&&&\textbf{Asian} & \textbf{Black} & \textbf{White} \\
\multirow{2}{*}{\textbf{Race}} & CLIP     & 6.26 & 14.49 & 80.36 & 73.59 & 82.93 & 74.60 & 81.23 \\
                      & PMC-CLIP & 8.12 & 7.15 & 81.23 & 76.04 & 83.27 & 77.09 & 81.87 \\ \midrule
&&&&&&\textbf{Female} & \textbf{Male} \\
\multirow{2}{*}{\textbf{Gender}} & CLIP    & 0.72 & 4.20 & 80.36 & 76.19 & 77.84 & 83.31 & \\
                      & PMC-CLIP & 3.66 & 11.07 & 81.23 & 76.42 & 78.24 & 84.54 & \\ \midrule
&&&&&&\textbf{Non-Hispanic} & \textbf{Hispanic} \\
\multirow{2}{*}{\textbf{Ethnicity}} & CLIP   & 16.32 & 18.25 & 80.36 & 76.98 & 80.49 & 76.11 & \\
                      & PMC-CLIP & 15.20 & 15.33 & 81.23 & 77.28 & 81.43 & 76.32 & \\ \midrule
&&&&&&\textbf{English} & \textbf{Spanish} & \textbf{Others} \\
\multirow{2}{*}{\textbf{Language}} & CLIP   & 21.78 & 22.28 & 80.36 & 69.76 & 80.70 & 72.16 & 73.71 \\
                      & PMC-CLIP & 10.31 & 22.28 & 81.23 & 70.53 & 81.73 & 76.70 & 71.08 \\ 
\bottomrule
\end{tabular}
\end{table*}

\end{document}